\documentclass[journal]{IEEEtran}
\usepackage{amsmath}
\usepackage{lineno}
\modulolinenumbers[5]
\usepackage{amssymb}
\usepackage{booktabs}
\usepackage{graphicx}
\usepackage{bm}
\usepackage{cite}
\usepackage{booktabs}

\usepackage[table,xcdraw]{xcolor}
\begin{document}
\title{MD Loss: Efficient Training of 3D Seismic Fault Segmentation Network under Sparse Labels by Weakening Anomaly Annotation}
\author{Yimin Dou, Kewen Li, Jianbing Zhu, Timing Li, Shaoquan Tan, Zongchao Huang
	\thanks{The corresponding author is Kewen Li. likw@upc.edu.cn}
	\thanks{Yimin Dou, Kewen Li, Zongchao Huang, College of computer science and technology, China University of Petroleum (East China) Qingdao, China.}
	\thanks{Timing Li, College of Intelligence and Computing, Tianjin University Tianjin, China.}
	\thanks{Jianbing Zhu, Shaoquan Tan, Shengli Oilfield Company, SINOPEC Dongying, China.}
	\thanks{This work was supported by grants from the National Natural Science Foundation of China (Major Program, No.51991365), and  the Natural Science Foundation of Shandong Province of China (ZR2021MF082).}
}

\maketitle
\begin{abstract}
Data-driven fault detection has been regarded as a 3D image segmentation task. The models trained from synthetic data are difficult to generalize in some surveys.
Recently, training 3D fault segmentation using sparse manual 2D slices is thought to yield promising results, but manual labeling has many false negative labels (abnormal annotations), which is detrimental to training and consequently to detection performance.
Motivated to train 3D fault segmentation networks under sparse 2D labels while suppressing false negative labels, we analyze the training process gradient and propose the Mask Dice (MD) loss.
Moreover, the fault is an edge feature, and current encoder-decoder architectures widely used for fault detection (e.g., U-shape network) are not conducive to edge representation.
Consequently, Fault-Net is proposed, which is designed for the characteristics of faults, employs high-resolution propagation features, and embeds Multi-Scale Compression Fusion block to fuse multi-scale information, which allows the edge information to be fully preserved during propagation and fusion, thus enabling advanced performance via few computational resources.
Experimental demonstrates that MD loss supports the inclusion of human experience in training and suppresses false negative labels therein, enabling baseline models to improve performance and generalize to more surveys.
Fault-Net is capable to provide a more stable and reliable interpretation of faults, it uses extremely low computational resources and inference is significantly faster than other models.
Our method indicates optimal performance in comparison with several mainstream methods.

%The Fault-Net parameter is only 0.42MB, support up to 528$^3$ (FP32) and 640$^3$ (FP16) size cuboid inference on 16GB RAM, its inference speed is significantly faster than other models. 
%Our approach employs fewer computational resources while providing more reliable and clearer interpretations of seismic faults.

\end{abstract}

% Note that keywords are not normally used for peerreview papers.

\IEEEpeerreviewmaketitle

\section{Introduction}
Fault detection is a crucial step for seismic structural interpretation, reservoir characterization and well placement, which determines that fault detection is an important topic in the field of oil-gas exploration. 
\subsection{Traditional approaches}
Traditional methods use theory of fault anisotropy \cite{ruger1997p,rueger1997using,crampin1984effective}, coherence \cite{marfurt19983,bahorich19953,gersztenkorn1999eigenstructure}, ant colony algorithms\cite{pedersen2002automatic,sun2011application,aqrawi2011improved} and edge detection algorithms \cite{chopra2014seismic,al2013fault,aqrawi2014adaptive} etc. to interpret faults, but these become problematic when undesired noises have a wavenumber spectrum similar to that of the image itself, and can be computationally intensive. 

\subsection{Deep learning approaches}
Some recent work views this as a deep learning based image segmentation task \cite{guo2018new,araya2017automated,wu2018convolutional,xiong2018seismic, wu2019faultseg3d, liu2020common, qi2020comparing, wei2022seismic, wu2019faultnet3d, feng2021uncertainty, gao2021fault, gao2022automatic,di2020accelerating, li2019seismic, cunha2020seismic, yan2021improving, wang2020distilling, an2021deep,dou2021seismic}.

The first applications to seismic fault segmentation were 2D networks\cite{guo2018new,araya2017automated,wu2018convolutional} etc., where the inputs and outputs are 2D slices. However, if 2D networks are predicted along the inline direction, the stitched 3D volumes will not be smooth in the crossline and timeline directions, which is a serious drawback, so researchers have gradually focused on 3D fault segmentation.

Xiong et al. extracted 2D slices of size 24 × 24 along inline, crossline and timeline for the centroid $O$ in a 3D seismic cube, and these three slices were used as a basis for predicting whether its centre was a fault or not\cite{xiong2018seismic}. In 3D seismic data, this approach clearly leads to redundant calculations and is sensitive to noise, one of the motivations for doing so is to obtain 3D labels, and obtaining accurate 3D fault labels has been a thorny problem plaguing the field.
\subsubsection{Based on synthetic data}
Wu et al. proposed FaultSeg3D, which uses synthetic data to train 3D UNet \cite{wu2019faultseg3d}. Synthetic data avoids the series of problems associated with manual 3D labels, and this pioneering work greatly improves the performance of seismic fault segmentation tasks.
Since then, most of the tasks for 3D fault segmentation have used the synthetic data provided by Wu. Liu et al. used ResUNet and synthetic data for training and obtained better results than UNet\cite{liu2020common}.
Qi and Wu et al. compared the performance of fault detection using synthetic data and traditional statistical methods \cite{qi2020comparing}. Wei et al. used focal loss to improve fault segmentation with synthetic data \cite{wei2022seismic}. Wu et al. used a single neural network to simultaneously predict the probabilities, strikes and dips of faults \cite{wu2019faultnet3d}. Feng et al. improved FaultSeg3D by adding dropout and Bayesian inference \cite{feng2021uncertainty}. Gao obtained very promising results using an improved Nested UNet (Nested Residual UNet, NRU) and synthetic data \cite{gao2021fault}, and subsequently he proposed MACNN with more reliable results\cite{gao2022automatic}. All these methods are trained by synthetic data, but there is a serious drawback of using synthetic data, deep learning models require a large amount of data, and the limited synthetic data cannot guarantee that the trained models can generalize stably under any geological conditions of fault development, which many researchers are aware of and have carried out some work.

\subsubsection{Based on field data}
Di et al. labeled some 2D slices in Opunake-3D field data, trained a 2D segmentation network, and then used the network to predict the remaining unlabeled slices, and trained a 3D fault segmentation network by stitching the 3D labels from the predicted 2D labels \cite{di2020accelerating}, similar to this is the approach of li et al \cite{li2019seismic}. Although this method introduced field data, the 3D labels stitched from the 2D labels are not smooth and will affect the performance of the model.
Similarly, to address the discrepancy between synthetic seismic data and field data, Cunha et al. introduced transfer learning, using 2D synthetic data as a pre-trained model that was fine-tuned in the manually labeled 2D data \cite{cunha2020seismic}.
Yan used a model trained with 3D synthetic data as a pre-trained model, and then fine-tuned the 3D labels obtained using the RANSAC method \cite{yan2021improving}, but the performance of the method was unsatisfactory due to the limitation of RANSAC fault detection accuracy.
Wang used CNNs trained with manually labeled data and CNNs trained with synthetic data as teacher networks to train student CNNs, respectively \cite{wang2020distilling}. An et al. proposed a manually labeled dataset to compensate for the limitations of the synthetic dataset \cite{an2021deep}.

In our previous work, we proposed $\lambda$-BCE loss and AAM, which can complete end-to-end training of 3D fault detection networks using a few 2D labels \cite{dou2021seismic}, and only need to manually label 2D slices accounting for 3.3\% of 3D data to achieve similar performance as using all labels, and it mainly works on the principle of gradient cropping of unlabeled regions. The method reduces the workload and difficulty of labeling and greatly extends the application of CNNs in fault recognition tasks, allowing it to be generalized to more field investigations. However, the work still has limitations in that BCE loss is too sensitive to incorrect labeling, while manual labeling is difficult to guarantee its accuracy, and its training effect is highly dependent on label quality.

The above work shows that although numerous researchers and groups expect to incorporate field data's into the training of fault segmentation networks, there are two main challenges: the first is the huge workload of labeling 3D faults. The second is that it is difficult to ensure the reliability of manual labeling, and incorrect labels can affect model performance.

\subsection{Our approach}

While $\lambda$-BCE greatly reduces the annotation effort by introducing 2D labels into the training of 3D models, it is overly dependent on the quality of annotation. Manual annotation is prone to produce a large number of false negative labels (FNL).

When manually labeling faults, it is easy to observe faults perpendicular to the slice due to the anisotropy of the faults, whereas faults parallel or nearly parallel to the slice become difficult, and when the angle between the fault and the slice exists, it will show some 'width' on the slice image, and as the angle becomes increasingly parallel to the slice, this 'width' will increase, and it is also difficult for the human to observe these 'widths' (Fig. \ref{fig4}).
Therefore, in the experiments of this work of $\lambda$-BCE \cite{dou2021seismic}, the performance of labeling only inline slices is better than labeling both inline and crossline, while the control group labeling only crossline shows the worst performance. The main reason for this is that the faults are mostly parallel to the crossline, which has many fault planes that are difficult to be observed by the human eye, which leads to many FNLs, and the loss based on cross entropy is extremely sensitive to FNL, which in turn misleads the gradient descent process of training.
Since most of the faults are perpendicular to the inline slices, the inline should be labeled when manually labeling the 3D seismic images, but there are still many surveys with criss-cross faults, it has many faults parallel to the inline or with 'width'. This results in many FNLs during the labeling process, which can seriously jeopardize the training process. Therefore, we aim to reduce this undesirable effect by a new loss function that needs to satisfy both the training under sparse 2D labeling and the suppression effect on FNL in order to address both challenges of introducing field data into the training of 3D seismic fault detection networks at present.

Segmentation loss functions are divided into two main categories, distribution-based and region-based \cite{ma2020segmentation}. Distribution-based loss aims to minimize the difference between two distributions, and the common practice is to treat each pixel as a sample\cite{wu2019faultseg3d, dou2021seismic}, i.e., calculate the cross-entropy loss for each pixel and then average it. When using distribution-based loss for training under sparse labels, the general practice is to ignore the gradient of unlabeled pixels \cite{cciccek20163d,dou2021seismic}. Region-based loss aims to minimize the mismatch or maximize the overlap region between ground truth and predicted segmentation results\cite{milletari2016v,rahman2016optimizing}, which limits the application of it to training on sparsely labeled data, thus all the current training under 3D sparse labels are distribution-based methods\cite{cciccek20163d,dou2021seismic}.

In this work, we analyze two types of loss in the form of gradients and find that region-based loss can weaken the effect of FNL and is preferred for training in seismic fault detection. 
To extend this type of loss to work under sparse labels, we propose Mask Dice loss (MD loss) and demonstrate mathematically and experimentally the suppression of FNL by this loss, while supporting the training of 2D labels on 3D networks. This is the first reported region-based segmentation loss that can be trained with sparse labels.

Faults can be characterized by edge features, and some works have achieved promising results in detecting faults by image gradient operators (edges).\cite{chopra2014seismic,al2013fault,aqrawi2014adaptive}.
While current CNN-based fault detection uses encoder-decoder structures (U-shape, such as UNet, UNet++, etc.) \cite{wu2019faultseg3d,gao2021fault,cunha2020seismic,liu2020common}, these structures disfavor the characterization of edge features. 
The encoding process (multiple downsampling) essentially transforms detailed information such as edges into high-level semantic information, which can achieve promising results in medical image segmentation tasks, but Long et al. mentioned that CNN encoding leads to degradation of image details (edges, etc.) \cite{shelhamer2016fully}, so these methods may lead to loss of edge information in seismic images. To prevent information degradation caused by encoding, these structures use larger widths.
In addition, the theory-driven approach demonstrates that for edge features such as faults only limited neighborhoods around the fault are needed for their efficient identification, such as coherence and sobel \cite{marfurt19983,bahorich19953,gersztenkorn1999eigenstructure,chopra2014seismic,al2013fault,aqrawi2014adaptive}. Whereas U-shape networks have more layers, these structures try to establish full image pixel or voxel correlations, i.e., receptive field covering the whole image, mainly because U-shape was originally designed to handle medical images\cite{ronneberger2015u, zhou2018unet++}, and for seismic fault detection, excessive layers are unnecessary, leading to parameter redundancy and even overfitting.

We propose Fault-Net, which differs from the encoder-decoder structure by using a novel  CNN structural paradigm that represents features in high resolution \cite{wang2020deep} and redesigned for the characteristics of seismic fault. 
Fault-Net has no coding process and the edge information is less degraded during propagation. In order to preserve the edge information in feature fusion as well, we embed the lightweight feature fusion block Multi-Scale Compression Fusion (MCF), which decouples the convolution process into feature selection and channel fusion to prevent image details from being weakened during fusion. Fault-Net fully preserves the edge information of the faults, so that it is not necessary to use a wider and deeper structure to guarantee the reliability of the results as in other networks, and we can achieve more promising results with only a very few computational resources.
Parameters and FLOPs of Fault-Net with tiny (Parameter: 0.42MB, FLOPs: 16.12G/128$^3$), inference for Netherlands F3 data of size $128\times 512\times 384$ on GPU takes only 0.55s, on CPU 4.48s. Support 528$^3$ (FP32) and 640$^3$ (FP16) size cuboid inference on 16G RAM, it can handle most field data without cubing. Its speed is significantly faster than other models. Saves computing resources while providing state-of-the-art fault detection performance.

In short, we propose MD loss, which is the first reported region-based segmentation loss function under sparse labels, and it can solve the prevalent FNL problem in manually labeled seismic image. Based on the characteristics of seismic faults, Fault-Net is proposed, which can obtain very high quality fault segmentation results with a few parameters and computational effort.

\section{Approach}
In this section, we first introduce and discuss MD loss with a view to addressing the two main challenges in training seismic fault detection networks, and then introduce the Fault-Net.

\subsection{Mask Dice Loss}

\begin{figure}[htb]
	\includegraphics[scale=0.42]{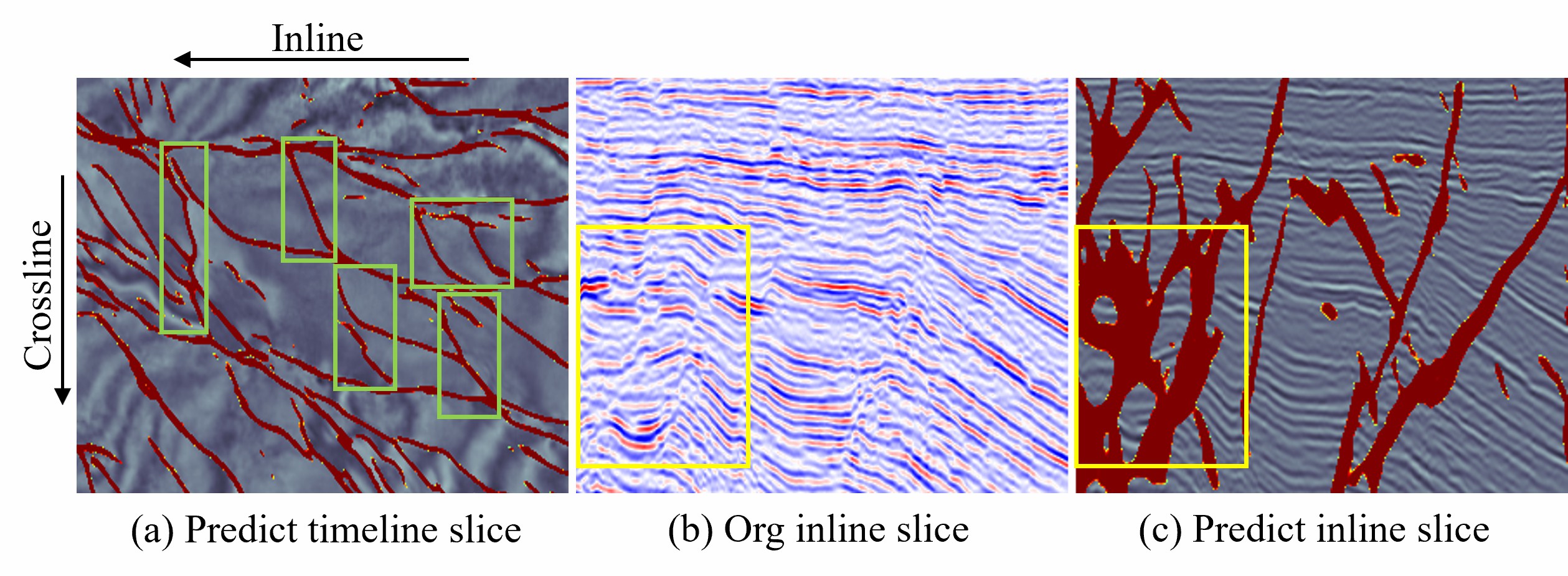}
	\centering\caption{Most of the faults in the seismic image are perpendicular to the inline direction of the slice, so when labelling faults, the slice is generally made along the inline direction, but some data have crisscross faults, which may result in the slice being parallel to the fault (green box), and slices that are parallel to the faults are difficult to observe with the human eye (yellow box), which is one of the reasons for FNL. There are also certain faults that are very difficult to annotation manually even if they are not parallel to the slice.}
	\label{fig4}
\end{figure}

The peculiarities of the fault structure in seismic image make it almost impossible to label them completely in 3D, so in our previous work we proposed to use a combination of sigmoid and binary cross-entropy (BCE loss) for training seismic image under sparse labels\cite{dou2021seismic}. The method significantly reduces the annotation effort, but the slices of seismic image have many faults that are not observable to the human eye, as shown in Fig. \ref{fig4}, which leads to massive FNL and thus adversely affects the training process.

However, the distribution-based loss is very sensitive to the FNL. BCE loss is represented by Equation (\ref{bce}).
\begin{equation}
	\mathcal{L}_{\text{bce}}(\hat{y},y)=-\frac{1}{n}\sum_{i}^{n}(y_i\log{\hat{y}_i}+(1-y_i)\log{(1-\hat{y}_i)})\label{bce}
\end{equation}

where $\hat{y}_i=\text{sigmoid}(\hat{x}_i)$, $\hat{x}_i$ is the output of the neural network. Its expression and gradient form are defined as Equation (\ref{sigmoid}).
\begin{equation}
	\hat{y}_i=\frac{1}{e^{\hat{x}_i}+1}, 
	\frac{\partial \hat{y}_i}{\partial \hat{x}_i}=\hat{y}_i(1-\hat{y}_i)\label{sigmoid}
\end{equation}
The partial derivative of the loss function to the output of the neural network is expressed as,

\begin{equation}
	\begin{aligned}
		\frac{\partial\mathcal{L}_{\text{bce}}(\hat{y},y)}{\partial \hat{x}_k}&
		=\frac{\partial\mathcal{L}_{\text{bce}}(\hat{y},y)}{\partial \hat{y}_k}\cdot\frac{\partial \hat{y}_k}{\partial \hat{x}_k}  \\
		&
		=-(\frac{y_k}{\hat{y}_k}+\frac{1-y_k}{\hat{y}_k-1})\cdot(\hat{y}_k(1-\hat{y}_k))\\
		&
		= \hat{y}_k-y_k
	\end{aligned}
\end{equation}
in the case of FNL, that is, $y_k=0$, in the later stages of training, $\hat{y}_k$ trend to 1. The gradients caused by FNL at this point are very large and have a negative impact on training. It also causes the gradient of the model to drop in the opposite direction in the early stages of training. 

Distribution-based losses does not distinguish between foreground and background in the segmentation task. When the current voxel label is for the background, the resulting gradient is the Euclidean distance between the predicted and labeled values, i.e. the gradient is proportional to the distance between the two, and thus FNL causes a significant adverse effect on training. 

The representative region-based segmentation loss is dice loss, it represented by Equation (\ref{dice1}). 
\begin{equation}
	\mathcal{L}_{\text{dice}}=1-\frac{2|\hat{Y}\cap Y|}{|\hat{Y}|+|Y|}\label{dice1}
\end{equation}
Its derivable form is Equation (\ref{dice2}).

\begin{equation}
	\mathcal{L}_{\text{dice}}(\hat{y},y)=1-\frac{\sum_{i}^{n}\hat{y}_iy_i}{\sum_{i}^{n}(\frac{1}{2}\hat{y}_i+\frac{1}{2}y_i)}\label{dice2}
\end{equation}
Where $\hat{y}_i=\text{sigmoid}(\hat{x}_i)$, the partial derivative of the loss function to the output of the neural network is expressed as,

\begin{equation}
	\begin{aligned}
	\frac{\partial\mathcal{L}_{\text{dice}}(\hat{y},y)}{\partial \hat{x}_k}&=\frac{\partial\mathcal{L}_{\text{dice}}(\hat{y}_i,y_i)}{\partial \hat{y}_k}\cdot\frac{\partial \hat{y}_k}{\partial\hat{x}_k}  \\&=
	\frac{\frac{1}{2}\alpha_\text{dice}-y_k\beta_\text{dice} -\frac{1}{2}y_k^2}{(\frac{1}{2}\hat{y}_k+\frac{1}{2}y_k+\beta_\text{dice})^2}\cdot\frac{\partial \hat{y}_k}{\partial\hat{x}_k}  \\
	\end{aligned}
\end{equation}
where, $\alpha_\text{dice}= \sum_{i,i\ne k}^{n}\hat{y}_iy_i$, $\beta_\text{dice}= \sum_{i,i\ne k}^{n}(\frac{1}{2}\hat{y}_i+\frac{1}{2}y_i)$. If the current voxel label is background, let $y_k=0$.

\begin{equation}
	\frac{\partial\mathcal{L}_{\text{dice}}(\hat{y},y)}{\partial \hat{x}_k}=\frac{ \alpha_\text{dice}}{2(\frac{1}{2}\hat{y}_k+\beta_\text{dice})^2}\cdot\frac{\partial \hat{y}_k}{\partial\hat{x}_k}  \label{7}
\end{equation}

Approximate $\frac{ \alpha_\text{dice}}{2(\frac{1}{2}\hat{y}_k+\beta_\text{dice})^2}$ as a constant, so when the current label is background, its gradient is related to $\frac{\partial \hat{y}_k}{\partial \hat{x}_k}$. We refer to this as the dice gradient coefficient, find its derivative as.
\begin{equation}
	\frac{\partial \hat{y}_k}{\partial \hat{x}_k\hat{y}_k}=1-2\hat{y}_k\label{8}
\end{equation}
From Equation (\ref{8}) when $\hat{y}_k= 0.5$, the dice gradient coefficient achieves the maximum value (at the initial of training), i.e. the dice gradient coefficient due to background is the largest, and as training progresses, the gradient coefficient due to background decreases in square steps when the predicted values tend to be in a certain category. Therefore if the labels are background, a larger gradient will not be triggered even if the predicted values are too different from the actual labels. 
Assuming that the voxel is FNL, in the later stages of training, $\hat{y}_k$ should tend to 1, then,
\begin{equation}
	\lim_{\hat{y}_k\to1} \frac{ \alpha_\text{dice}}{2(\frac{1}{2}\hat{y}_k+\beta_\text{dice})^2}\cdot \hat{y}_k(1-\hat{y}_k) = 0 \label{dice3}
\end{equation}
from Equation (\ref{dice3}), dice loss causes little effect from FNL in the late training period.
However, at the initial of the training period, $\hat{y}_k$ tends to 0.5, when the effect of FNL on dice loss is also greater.

We hope to introduce the good properties of dice loss into the training of sparse labels and to minimize the adverse effects of FNL in the early stage of model training. 

Firstly, in order to migrate the dice loss to train the 3D segmentation network under sparse labels, we proposed Mask Dice loss (MD loss).

\begin{figure}[htb]
	\includegraphics[scale=0.4]{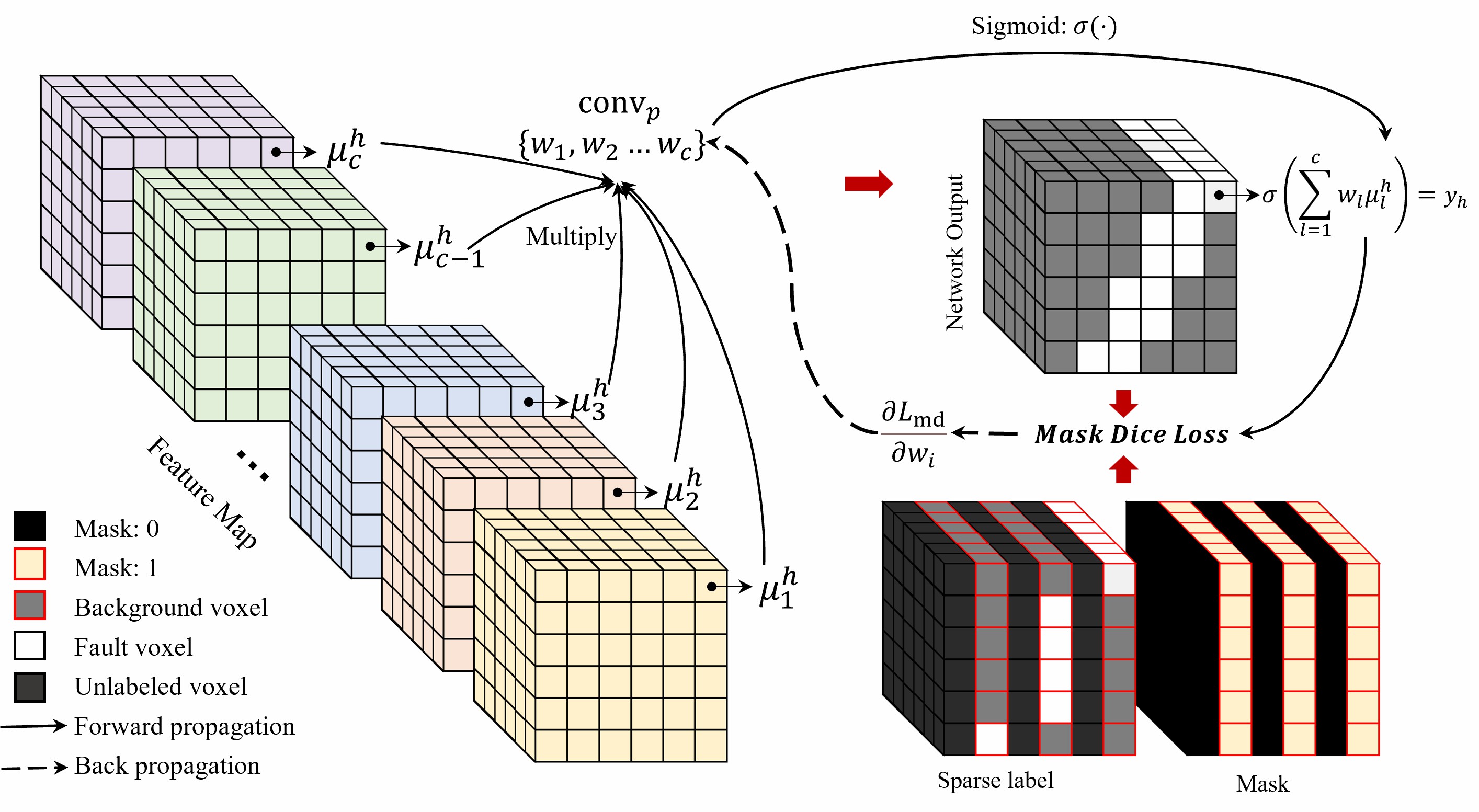}
	\centering\caption{The values $\mu$ at the same position in different channels on the feature map share the parameter $\{w_1,w_1,...w_c\}$ of conv$_p$. The key to using sparse labels is to find the valid gradient about $\{w_1,w_1,...w_c\}$ using the loss function, and we obtain the MD loss by incorporating the mask mechanism in the dice loss and deriving its differentiable form.}
	\label{fig5}
\end{figure}
Fig. \ref{fig5} expresses the process of MD loss using 2D slices to train a 3D network, and next we detail the derivation of MD loss and how it works.

Generally, the prediction layer of the segmentation network consists of a $1\times1$ convolution (conv$_p$) and a sigmoid activation function ($\sigma(\cdot)$), the last layer of feature maps extracted by backbone shares the parameters $\{w_1,w_1,...w_c\}$ of conv$_p$. 
The forward propagation process is as follows. The set of values of the same location $h$ of $c$ channels in the Feature map is denoted as $\{\mu_1^h,\mu_2^h...\mu_c^h\}$, and its weighting process by conv$_p$ is denoted as $\sum_{l=1}^{c}w_l\mu_l^h$, the predicted result at position $h$ is denoted as Equation (\ref{predict}).
\begin{equation}
y_h = \sigma(\sum_{l=1}^{c}w_l\mu_l^h)\label{predict}
\end{equation}
Since the convolution is a shared mechanism, positions other than $h$ follow the same expressions and convolution parameters. Therefore, in back propagation, we only need to find the valid gradient for $\{w_1,w_1,...w_c\}$. The output of the network is a 3D volume and the labels are 2D slices. The distribution-based loss treats each voxel as a separate sample, so their average can be propagated back as a valid gradient by calculating the gradient at the corresponding position of the 2D label \cite{dou2021seismic}. However, it is difficult to perform the same operation with region-based loss, so we introduce the Mask mechanism, where the label consists of two parts: 2D labels and masks.

Region-based loss aims to minimize the mismatch or maximize the overlap region between ground truth $Y$ and predicted results $\hat{Y}$, Its original expression is Equation (\ref{dice1}). We want to calculate the overlap only for the regions corresponding to 2D labels, and Equation  (\ref{dice1}) becomes Equation  (\ref{dice_md}, \ref{mdloss2}) after introducing Mask $\mathcal{M}$.
\begin{equation}
	\mathcal{L}_{\text{dice}}=1-\frac{2|\mathcal{M}(\hat{Y})\cap \mathcal{M}(Y)|}{|\mathcal{M}(\hat{Y})|+|\mathcal{M}(Y)|}\label{dice_md}
\end{equation}
\begin{equation}
	\begin{aligned}
		\mathcal{M}_i=&\begin{cases}
			0 &\mbox{NonLabeled} \\
			1 &\mbox{Labeled}
		\end{cases}
	\end{aligned}\label{mdloss2}
\end{equation}
Its differentiable form is the MD loss, Equation (\ref{10}).
\begin{equation}
	\begin{aligned}
		\mathcal{L}_{\text{md}}(\hat{y},y)=&1-\frac{\sum_{i}^{n}\mathcal{M}_i\hat{y}_iy_i}{\sum_{i}^{n}\mathcal{M}_i(\frac{1}{2}\hat{y}_i+\frac{1}{2}y_i)}\label{10}
	\end{aligned}
\end{equation}
The value domain of the MD loss is $[0,1]$, which is the combined loss of the predicted value of the sparse region corresponding to the 2D label position and the label, so its partial derivative $ \frac{\partial\mathcal{L}_\text{md}}{\partial w}$ for  $\{w_1,w_1,...w_c\}$ can be used as the valid gradient.

It is worth emphasizing that although our method uses 2D slice labels for training, both the input and output of the network are 3D volumes, our network does not use individual 2D slices during training. Labels are inserted into 3D volumes with 2D labels, where fault regions are labeled as 1, non-fault (background) regions are labeled as 0, and regions without labels are marked with '-1' in order to generate masks to provide MD loss (you can see the label structure in Fig. \ref{label}).
The results are also fundamentally different from the 2D network in that our output is smooth in all directions. In contrast, the 2D network input and output are 2D slices, and if the prediction is performed along the inline, the spliced 3D volumes are not smooth in the crossline and timeline directions.

\begin{figure}[htb]
	\includegraphics[scale=0.28]{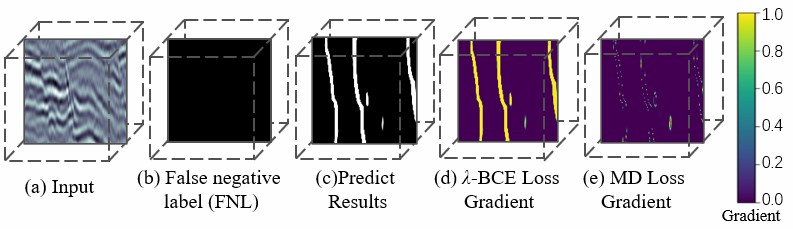}
	\centering\caption{The model has the ability to segment faults, which can be regarded as the later stage of model training. (d) shows that the  $\lambda$-BCE is very sensitive to FNL labels and (e) shows that the response of MD loss to FNL is very low in the late training period.}
	\label{fig3}
\end{figure}

Fig. \ref{fig3} shows the gradient responses of $\lambda$-BCE\cite{dou2021seismic} and Mask Dice in the later stages of training, when the labels are false negatives.

Then, to further attenuate the detrimental effect of FNL on training (mainly the initial stage of training), we introduce the the coefficient $\gamma$, Equation (\ref{mdloss1}).
\begin{equation}
	\begin{aligned}
			\mathcal{L}_{\text{md}}(\hat{y},y)=&1-\frac{\sum_{i}^{n}\mathcal{M}_i\hat{y}_iy_i}{\sum_{i}^{n}\mathcal{M}_i((1-\gamma)\hat{y}_i+\gamma y_i)} 
	\end{aligned}\label{mdloss1}
\end{equation}
where $\gamma\in[0.5,1)$, it is the weight coefficient of the fault and background,  record it as MD$^\gamma$ loss, when $\gamma=0.5$, it is the original MD loss. We analyses the role it plays in the training by deriving this loss function.

\begin{equation}
	\begin{aligned}
		\frac{\partial\mathcal{L}_{\text{md}}(\hat{y},y)}{\partial x_k}&=\frac{\partial\mathcal{L}_{\text{md}}(\hat{y}_i,y_i)}{\partial \hat{y}_k}\cdot\frac{\partial \hat{y}_k}{\partial \hat{x}_k}  \\&=
		\frac{(1-\gamma)\alpha_\text{md} -\gamma y_k^2-y_k\beta_\text{md}}{((1-\gamma) \hat{y}_k+\gamma y_k+\beta_\text{md})^2}\cdot\frac{\partial \hat{y}_k}{\partial \hat{x}_k} \\
	\end{aligned}
\end{equation}
where, $\alpha_\text{md}= \sum_{i,i\ne k}^{n}\mathcal{M}_i\hat{y}_iy_i$, $\beta_\text{md}= \sum_{i,i\ne k}^{n}\mathcal{M}_i((1-\gamma)\hat{y}_i+\gamma y_i)$. If the current voxel is background, i.e. $y_k = 0$.
\begin{equation}
	\frac{\partial\mathcal{L}_{\text{md}}(\hat{y},y)}{\partial x_k}=(1-\gamma) \frac{ \hat{y}_k(1-\hat{y}_k)\alpha_\text{md}}{((1-\gamma)\hat{y}_k+\beta_\text{md})^2}\label{mdgrid}
\end{equation}
Equation (\ref{mdgrid}) shows that the gradient due to MD loss in the case of the current being a background voxel can be controlled by $\gamma$. Increasing $\gamma$ weakens the sensitivity of the model to the background in the early stages of training, making it more inclined to predict the foreground and further weakening the effect of FNL in the later stages of training.

In this section, we discuss the drawbacks of distribution-based losses for seismic fault segmentation tasks and the advantages of region-based losses. In order to migrate the dice loss to 3D segmentation tasks under sparse labels, we proposed MD loss, and theoretically demonstrating its suppressive effect on FNL. MD loss can train 3D fault segmentation networks using 2D slice labels and weakening anomalous annotations.

\subsection{Fault-Net}
Fault-Net propagates features of different scales forward in parallel to reduce degradation of image details by maintaining high resolution. The MCF block is embedded to preserve edge information when fusing features of different scales by decoupling the fusion process of convolution. It enables the network to obtain reliable results using less computational resources.

\subsubsection{Fault-Net Structure}
\begin{figure}[htb]
	\includegraphics[scale=0.225]{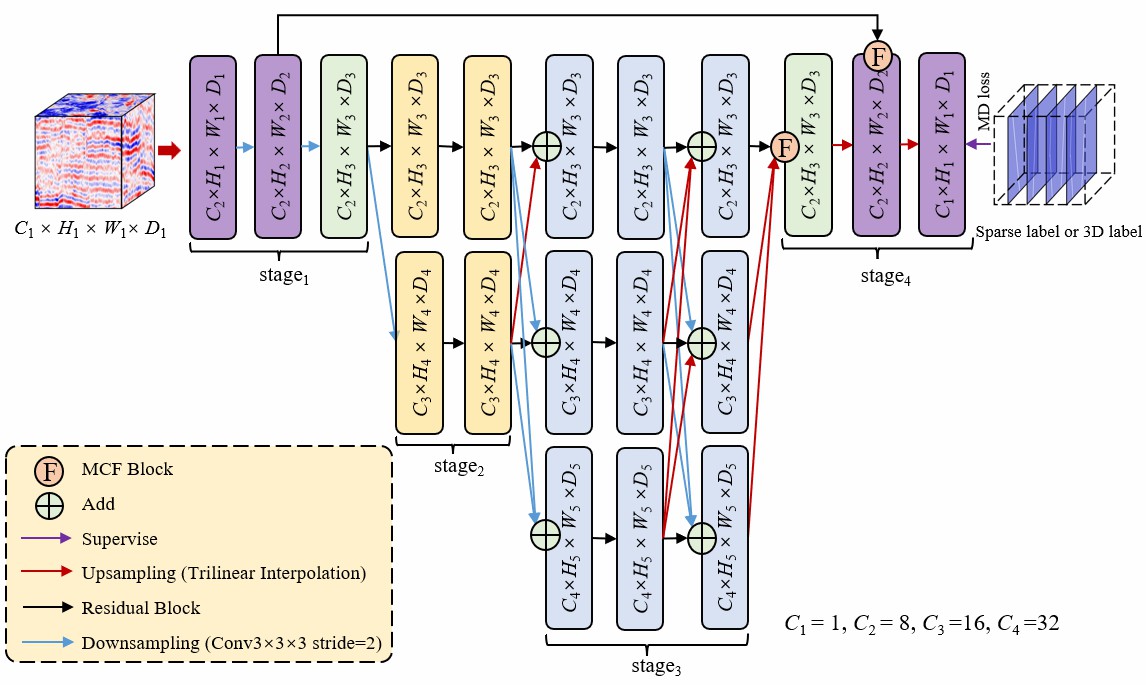}
	\centering\caption{Fault-Net Structure. Although we also use low-resolution features, not by concatenation, but by adding two low-resolution branches while preserving the high-resolution features. Thus Fault-Net always keeps the high-resolution propagation features after two downsampling, which allows the edges to be fully preserved. Also, when it is necessary to fuse features of different resolutions (multi-scales), we embed the proposed MCF block so that the edge features are not distorted during fusion.}
	\label{fig2}
\end{figure}
We use resblock as the base unit of the model  and the model has four stages. First, the model is first downsampled twice using two convolutions with a stride of 2. Downsampling using the convolution operation maximises the retention of the original features of the data. Secondly, downsampling is again performed by convolution, but while retaining the high resolution features, i.e. adding a branch of low resolution features that propagate the two resolutions forward in parallel. Third, a low-resolution branch is added, which is obtained by fusing the downsampled features of the other two branches, while the other two branches also perform feature fusion, and at the end of this stage, the features of the three branches are fused with each other. Feature fusion allows the three resolution features of the model to interact with each other, so that the high resolution features can obtain the semantic information of the low resolution features, and the low resolution features can obtain the image detail information of the high resolution features. Finally, the features of the two low-resolution branches are upsampled and fused with the high-resolution features, and the final segmentation result is obtained after two upsampling and residual.

In general, traditional CNN models double the width of the features for each down-sampling to attenuate the information loss due to sampling, whereas Fault-Net does not require an excessive width because it maintains edge feature when propagating forward, with three branches of 8, 16 and 32 widths respectively.
Faults in seismic image exist in the form of edge features, which are a low-level feature for images and which are easily characterised, so the general CNN model overlaying too many layers is redundant for the task of fault segmentation of seismic image. Therefore Fault-Net uses a shallow layer structure, as shown in Fig. \ref{fig2}.

\subsubsection{Multi-Scale Compression  Fusion Block}
In $\text{stage}_4$ of Fig. \ref{fig2}, so as to obtain high quality segmentation results, the three scales of features need to be fused, and in order to ensure the effectiveness of the fusion and reduce parameter redundancy, the width of the fused features should be the same as the width of the individual features before fusion, and there are two general ways of fusing features. One is to upsample the low-resolution features and add them to the high-resolution features \cite{shelhamer2016fully}, and the other is to fuse by convolution \cite{ronneberger2015u}. The first approach does not require too much computation, but feature addition results in a significant loss of information. 
The second method uses convolutional fusion, which is essentially a weighting of channels, and these weights are fixed after training, but the features will differ with the input, so fixed weights in fusion will blur the edge feature and make it difficult to retain detailed information. 
Therefore, we propose the Multi-scale Compressive Fusion (MCF) block, which expresses the process of channel fusion and feature selection explicitly by decoupling the convolution, and uses the result of feature selection for channel compressive fusion to preserve more image details.
The three scale features are first compressed to the same width by $1\times1$ convolution, and then the two low resolution features are scaled up to the same size as the high resolution features by trilinear interpolation. We express the features before fusion as,
\begin{equation}
\begin{aligned}
	\mathcal{F}_1 = \text{F}^\text{conv}_1(\mathcal{F}_1';1,1,C_2,C_2)\\
	\mathcal{F}_2 = \text{F}^\text{conv}_2(\mathcal{F}_2';1,1,C_3,C_2)\\
	\mathcal{F}_3 = \text{F}^\text{conv}_3(\mathcal{F}_3';1,1,C_4,C_2)
\end{aligned}
\end{equation}
the three scales of features are concatenated and denoted as.
\begin{equation}
	\mathcal{F}^\text{cat} = \text{F}^\text{cat}(\mathcal{F}_\text{1},\mathcal{F}_\text{2},\mathcal{F}_\text{3})
\end{equation}
If convolution is used for feature fusion, the expression is Equation (\ref{conv1}).
\begin{equation}
	\mathcal{F}^\text{cat} = \text{F}^\text{add}(\mathcal{W}_\text{conv}\mathcal{F}^\text{cat})\label{conv1}
\end{equation}
We decouple this process into feature selection and channel fusion. Where  $\mathcal{W}_\text{conv}$ is the convolution weight, which is fixed after training, and $\mathcal{W}_\text{conv}\mathcal{F}^\text{cat}$ is the feature selection part of the convolution fusion. $\text{F}^\text{add}(\cdot)$ sums the weighted features and is the channel fusion part. We want the network to adjust $\mathcal{W}_\text{conv}$ adaptively to different inputs so as to retain more details.

\begin{figure}[htb]
	\includegraphics[scale=0.23]{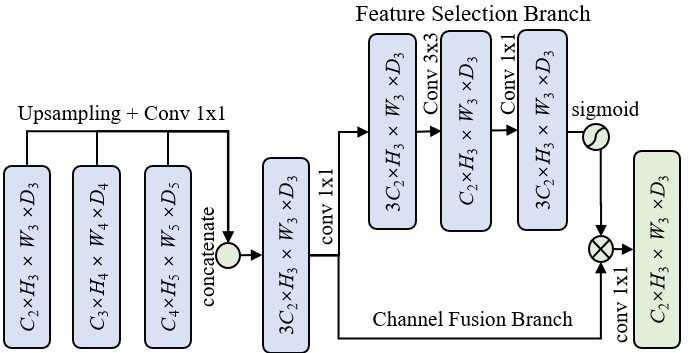}
	\centering\caption{Structure of MCF block. Different from the convolutional fusion with fixed weights, we decouple the convolutional fusion process into two branches of feature selection and channel fusion, separate the weighted part from the convolution, and generate weights adaptively according to the input to prevent the loss of edge information caused by the fusion.}
	\label{figmcf}
\end{figure}

We use two branches, one branch generates adaptive weights $\mathcal{W}_\text{conv}'$, which can be considered as the response of features in $\mathcal{F}^\text{cat}$ to the task, named feature selection branch, and the other branch compresses $\mathcal{F}^\text{cat}$ via $\mathcal{W}_\text{conv}'$, named channel fusion branch. The flow is shown in Fig. \ref{figmcf}. In the feature selection branch, the linear projection $\mathcal{F}^\text{cat}_1$ of $\mathcal{F}^\text{cat}$ is first computed, and by compressing the width and then recovering it, can be made to learn the sparse feature response of $\mathcal{F}^\text{cat}$ to the task, which is normalized with sigmoid to obtain $\mathcal{W}'_\text{conv}$. We express this process as Equation (\ref{18}).
\begin{equation}
	\begin{aligned}
	\mathcal{F}^\text{cat}_1&= \text{F}^\text{conv}_\text{4}(\mathcal{F}^\text{cat};1,1,3C_2,3C_2)\\
	\mathcal{W}'_\text{conv}=   \text{F}^\text{sigmoid}(\text{F}^\text{conv}_5&(\text{F}^\text{conv}_6(\mathcal{F}^\text{cat}_1;3,3,3C_2,C_2),1,1,C_2,3C_2))
	\end{aligned}	\label{18}
\end{equation}
All the features of  $\mathcal{F}_\text{cat}$ are retained in the channel fusion branch, multiplied with the feature response $\mathcal{W}'_\text{conv}$, and then compressed using convolution to obtain $\mathcal{F}^\text{press}$. The process is represented by Equation (\ref{19}). $\mathcal{F}^\text{press}$ is the compressed fused feature, which has the same size as the high resolution before fusion, and also contains the information of the three scale features.

\begin{equation}
	\mathcal{F}^\text{press} = \text{F}^\text{conv}_\text{press}(\mathcal{W}'_\text{conv}\odot \mathcal{F}^\text{cat},1,1,3C_2,C_2)\label{19}
\end{equation}

MCF is a lightweight feature fusion module that adaptively generates fusion weights based on different inputs, preserving more details. It is an important component of Fault-Net, which enables the edge information such as faults in the extracted high-resolution features to be fully preserved, prevents edge distortion during the fusion process. This component enables to output more reliable fault detection results by adding only a few parameters.

In this section, we describe Fault-Net and its design in detail. We integrate high-resolution representations into the network design based on the edge characteristics of seismic faults, i.e. three scales of features are propagated forward in parallel. In order to fully fuse the three scales of features for better decision making on the category of each voxel, we design and propose the MCF block, which decouples the convolution through an explicit representation feature selection process, stripping out redundant features while retaining valid features. 

Table \ref{t1} shows a comparison of Fault-Net with the mainstream lightweight network parametric quantities (Lite-HRNe, ShuffleNetV2, SqueezeNet  and MobilenetV3 \cite{mishra2020survey}), and a more comprehensive comparison of computational resources with previous work (FaultSeg i.e. UNet\cite{wu2019faultseg3d}), Nested UNet\cite{gao2021fault}, AAM-UNet\cite{dou2021seismic}).
Since Fault-Net takes into account the characteristics of faults, it requires significantly lower computational resources than other networks, and its performance has been experimentally demonstrated to be better than previous work.
\begin{table}[htb]
	\setlength\tabcolsep{1pt}
	\scriptsize
	\caption{Comparison of execution efficiency of different networks}
	\label{t1}
	\centering
	\begin{tabular}{cccccc}
		\hline
		\textit{\textbf{}}      & Parameters & \begin{tabular}[c]{@{}c@{}}FLOPs \\ (128$^3$)\end{tabular}  & \begin{tabular}[c]{@{}c@{}} Cuboid Size\\ (16G RAM)\end{tabular} & \begin{tabular}[c]{@{}c@{}}Infer Time \\ (128$^3$/GPU)\end{tabular} & \begin{tabular}[c]{@{}c@{}}Infer Time \\ (128$^3$/CPU)\end{tabular} \\ \hline
		Lite-HRNet-18 (2D)  & 1.10M      & -                                                                                                     & -                                                            & -                                                           & -                                                           \\
		
		ShuffleNetV2$\times$0.5 (2D)               & 1.37M      & -                                                                                                     & -                                                            & -                                                           & -                                                           \\
		SqueezeNet (2D)               & 1.25M      & -                                                                                                    & -                                                            & -                                                           & -                                                           \\
		MobilenetV3 small (2D)               & 2.54M      & -                                                                                                   & -                                                            & -                                                           & -                                                           \\
		UNet (3D)                   & 1.46M      & 136.46G                                                                                                 & 240$^3$                                                            & 0.21s                                                           & 2.44s                                                           \\
		Nested UNet (UNet++, 3D)                   & 6.55M      & 468.00.G                                                                                                 & 208$^3$                                                            & 0.43s                                                           & 3.12s                                                           \\
		AAM-UNet (3D)                   & 1.51M      & 138.80G                                                                                                 & 240$^3$                                                            & 0.22s                                                           & 2.58s                                                           \\
		Fault-Net (3D)                & 0.42M      & 16.12G                                                                                                   & 528$^3$                                                            & 0.13s                                                           & 0.82s                                                           \\
		Fault-Net (No MCF, 3D) & 0.39M      & 13.49G                                                                                                  & 528$^3$                                                            & 0.12s                                                           & 0.80s                                                           \\ \hline
	\end{tabular}
\end{table}

\section{Experiment}
\subsection{Data illustration}
Our field data comes from the Shengli Oilfield Branch of Sinopec, and thanks to Wu et al for published the synthetic seismic fault dataset \cite{wu2019faultseg3d}. There are significant differences in the numerical distribution of different seismic image, so in the experiment we have standardized and normalized all the data. 

The experiments were carried out on seismic images from three work areas, $\mathcal{A}$, $\mathcal{B}$ and $\mathcal{C}$, you can see the spatial distribution of the faults in these three work areas in Fig. \ref{exp1}, \ref{exp2}, \ref{exp3}.

The faults in  $\mathcal{A}$ and  $\mathcal{B}$ are mostly perpendicular to inline, and labeling is easy, so it can guarantee that most of the labels are accurate. Work area $\mathcal{C}$ has more crisscross faults, and there will be more FNL in labeling.

As mentioned above, manual labeling is inevitably inaccurate, and the employment of manual labels in the validation set may lead to errors in the quantification results.
Therefore, we use synthetic data as the validation set to ensure the reliability of the quantitative results, while using the mixture of field data and synthetic data to form the training set, MD loss support is trained in the presence of both 2D sparse labels (field data) and 3D labels (synthetic data), and the mixed data set ensures that the model can generalize robustly across more surveys, and more importantly, we can thus obtain the effect of FNL on different loss functions.

The synthetic data published by Wu are divided into 200 cuboids in train set and 20 in validation set, the size of each cuboid is 128$^3$. 
To enable the model to learn the fault features of seismic images at different scales, each synthetic data is upsampled to 256$^3$ or 384$^3$ and random cuboid of 128$^3$ is intercepted until the synthetic dataset reaches 600 training samples and 120 validation samples.

Work \cite{dou2021seismic} showed that labeling only inline slices (slices perpendicular to the main faults) gives the best performance, and adding crossline slices (slices parallel to the main faults) may be detrimental to the training of the model because faults parallel to the slices are not easily observed and manual labeling tends to cause significant FNL, although MD loss can attenuate the effect of FNL, it is mathematically does not completely avoid FNL, especially in the early stages of training. Another reason why the network can be trained effectively even if only one direction labeled is that the data augmentation method of random rotation is used for the samples in training, which allows slices labeled inline to be potentially rotated to crossline. We use the conclusions drawn from the experiments in work \cite{dou2021seismic} and try to avoid using slices parallel to the main faults in the field data (crossline) as manual labels during training. It was also shown in \cite{dou2021seismic} that labeling 3.3\% of the inline slices is the most efficient and the trained model has similar performance to using the full labeling of 3D, this is so because the adjacent slices in the 3D volume are extremely similar and adding redundant annotations do not affect the final performance too much, so we continue this labeling approach.

\begin{figure}
	\includegraphics[scale=0.42]{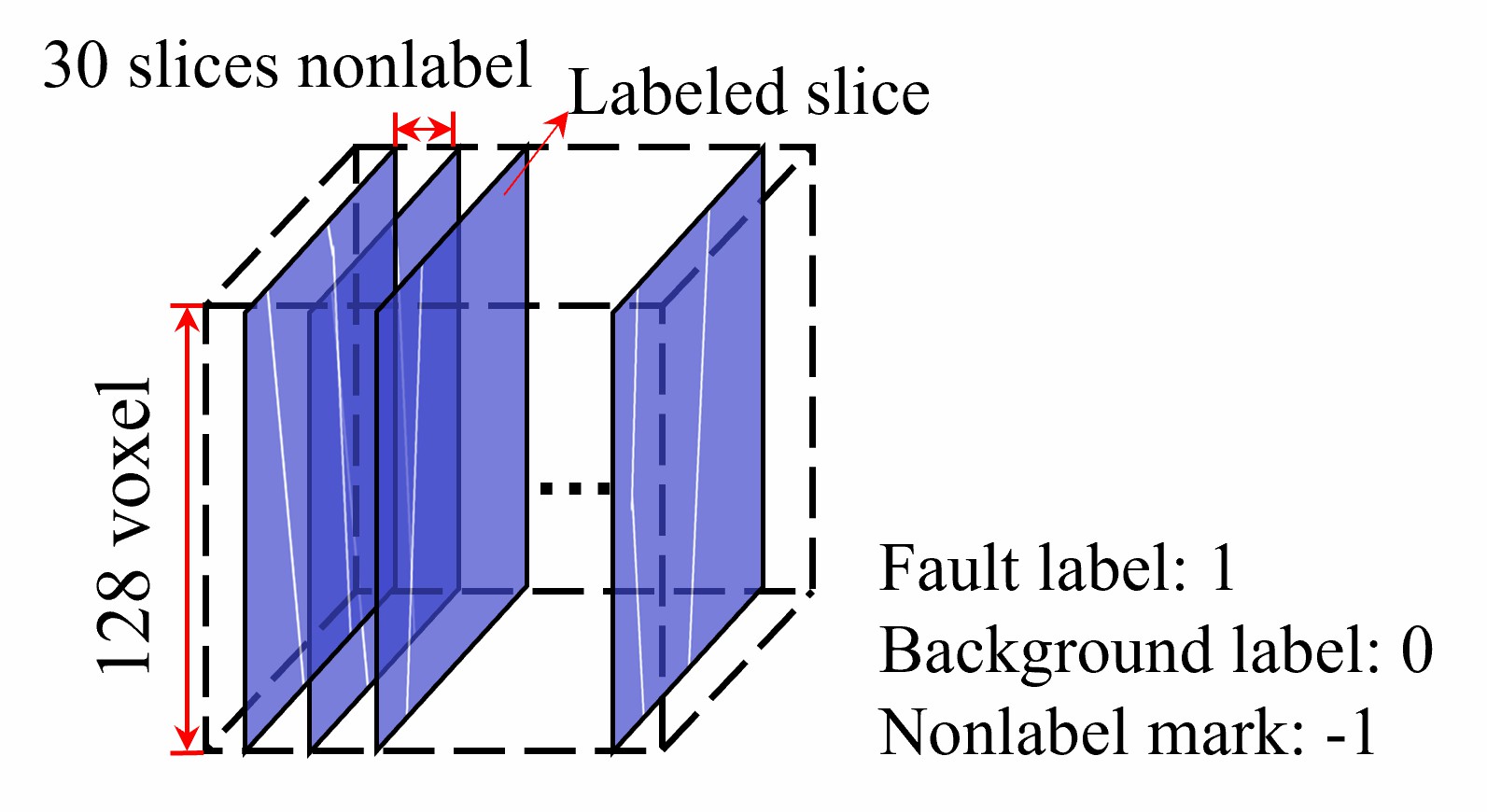}
	\centering\caption{Label structure. Sparse labels are used as supervisory information, where fault regions are labeled as 1, background regions are labeled as 0, and unlabeled regions are marked with '-1' to provide mask to MD loss.}
	\label{label}
\end{figure}

The labeling process is as follows. First, on the inline, label one slice every 30 slices. Second, slide sampling in three directions, the sliding stride is 35, and the size of each sample is $128\times128\times128$. When sampling, samples with fault voxels less than 128 are deleted. Finally, the redundant samples are randomly deleted, so that the number of samples to 300. There are more detailed labeling steps in  \cite{dou2021seismic}. Fig. \ref{label} illustrates the structure of the label.

We mixed field data with synthetic data for the experiment and it has four groups, the train set, validation set and test set are shown in Table \ref{t2}.
The experiments use six loss functions, MD$^{0.5}$, MD$^{0.6}$, MD$^{0.7}$, MD$^{0.8}$, MD$^{0.9}$ (represent $\gamma=0.5,0.6,0.7,0.8,0.9$ respectively) and $\lambda$-BCE loss. We use random rotation ($0^\circ,90^\circ,180^\circ,270^\circ$) for data augmentation.

\subsection{Experimental results and analysis}
We use IOU (Intersection Over Union) as the performance evaluation metric. The IOU is expressed by Equation (\ref{iou}).

\begin{equation}
	IOU=\frac{TP}{FP+TP+FN}\label{iou}
\end{equation}
Among them, TP (True Positive) is classified as a positive sample, in fact it is also a positive sample. FP (False Positive), is classified as a positive sample, but in fact it is a negative sample.
FN (False Negative) is classified as a negative sample, but in fact it is a positive sample. 
The output of the model is normalized to the probability $\hat{y}_i\in[0,1]$ through the sigmoid activation function, and $\hat{y}_i>0.5$ is regarded as a positive sample, $\hat{y}_i\leq0.5$ is regarded as a negative sample, so as to calculate the IOU.

The code is implemented by Pytorch 1.10.0, accelerated training by Apex\footnote{https://github.com/NVIDIA/apex}. The optimizer uses Adam with the learning rate of 0.001 and the batch size is 10. 
To conserve training time and improve convergence speed, in the quantization experiments, we first train a pre-training model (200 epochs) for each network using Dice loss and synthetic data, all experiments initialize the parameters with the pre-training model. The experiments are trained for 200 epochs (total 400 epochs), and each epoch is run once on the validation set to record the quantization metrics. Table \ref{t3} shows the best results for each set of experiments.

%Each model is trained for 500 epochs and run on the validation set at 100 steps per training, thus recording the quantified values, and Table \ref{t3} shows the best results for each set of experiments.

\begin{table}[htb]
	\caption{Train set and validation set partitioning}
	\scriptsize
	\label{t2}
	\centering
	\begin{tabular}{@{}cccccc@{}}
		\toprule
		& \textbf{Group-1}                  & \textbf{Group-2}                  & \textbf{Group-3}                  &\textbf{Group-4}                                         \\ \midrule
		Train & $\mathcal{A}, \mathcal{B},$ Synthetic & $\mathcal{A}, \mathcal{C},$ Synthetic & $\mathcal{B}, \mathcal{C},$ Synthetic & Synthetic  \\
		Validation  & Synthetic      & Synthetic      & Synthetic      & Synthetic                  \\ 
		Test  & $\mathcal{C}$      & $\mathcal{B}$      & $\mathcal{A}$      &  -                   \\ \bottomrule
	\end{tabular}
\end{table}

\begin{table}[htb]
	\setlength\tabcolsep{3pt}
	\scriptsize
	\caption{Quantitative experimental results}
	\label{t3}
	\centering
	\begin{tabular}{@{}ccccccc@{}}
		\toprule
		\multicolumn{7}{c}{\textbf{Group-1}}                                              \\ \midrule
		& MD$^{0.5}$  & MD$^{0.6}$  & MD$^{0.7}$  & MD$^{0.8}$  & MD$^{0.9}$  & $\lambda$-BCE  \\
		UNet     & 67.10       & 67.28       & 67.31       & 64.55       & 61.12       & 65.53    \\
		Nested UNet     & 66.57       & 67.20       & 67.25       & 64.22       & 61.01       & 65.61    \\
		Fault-Net & 67.01       & 67.32       & 67.24       & 64.69       & 61.83       & 65.50    \\ \midrule
		\multicolumn{7}{c}{\textbf{Group-2}}                                              \\ \midrule
		& MD$^{0.5}$  & MD$^{0.6}$ & MD$^{0.7}$ & MD$^{0.8}$ & MD$^{0.9}$ & $\lambda$-BCE  \\
		UNet     & 65.20       & 65.39       & 65.66       & 64.01       & 61.01       & 62.22    \\
		Nested UNet     & 65.66       & 65.50       & 66.21       & 63.90       & 61.45       & 62.88    \\
		Fault-Net & 65.89       & 66.04       & 66.18       & 63.73       & 61.29       & 62.87    \\ \midrule
		\multicolumn{7}{c}{\textbf{Group-3}}                                              \\ \midrule
		& MD$^{0.5}$  & MD$^{0.6}$  & MD$^{0.7}$ & MD$^{0.8}$ & MD$^{0.9}$  & $\lambda$-BCE  \\
		UNet     & 65.29       & 65.73       & 66.38       & 64.22       & 60.92       & 62.69   \\
		Nested UNet     & 65.99       & 65.98       & 66.31       & 64.10       & 61.00       & 63.11   \\
		Fault-Net & 65.87      & 66.19       & 66.56       & 64.31       & 61.48       & 63.32    \\ \midrule
		\multicolumn{7}{c}{\textbf{Group-4}}                                              \\ \midrule
		& MD$^{0.5}$  & MD$^{0.6}$  & MD$^{0.7}$ & MD$^{0.8}$ & MD$^{0.9}$ & $\lambda$-BCE \\
		UNet     & 66.90       & 67.18       & 67.27       & 64.19       & 62.10       & 66.35    \\
		Nested UNet     & 67.77       & 67.51       & 67.51       & 64.22       & 62.83       & 66.70    \\
		Fault-Net & 67.53       & 67.59       & 67.33       & 64.78       & 62.57       & 66.76    \\
		Fault-Net (No MCF) & 66.98       & 67.15      & 66.91       & 64.33       & 62.06      & 66.40    \\ \bottomrule
	\end{tabular}
\end{table}
\begin{figure}[htb]
	\includegraphics[scale=0.24]{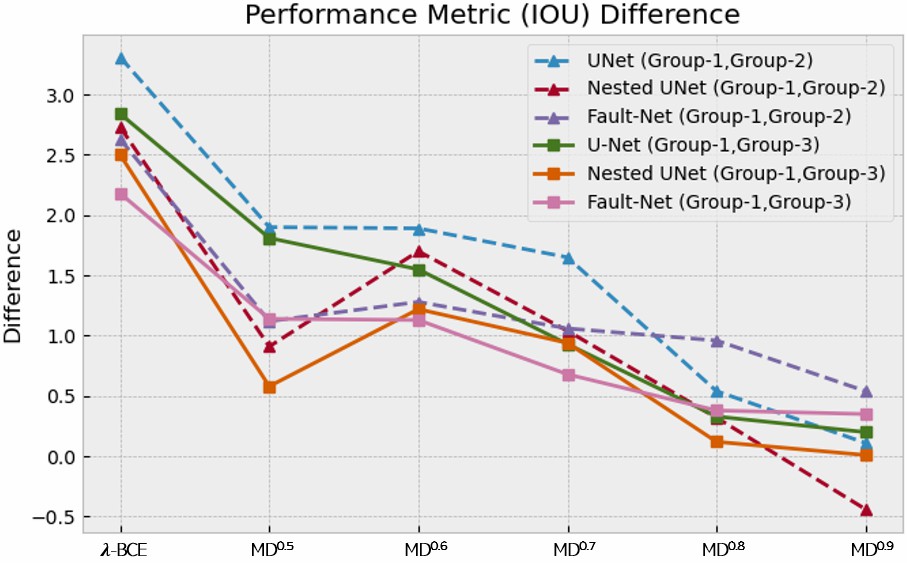}
	\centering\caption{The difference between Group 1 (less FNLs) and the other two groups (more FNLs) decreases with increasing $\gamma$ under MD loss. Overall $\gamma$-BCE shows the greatest performance difference. }
	\label{exp0}
\end{figure}
\begin{figure*}
	\includegraphics[scale=0.55]{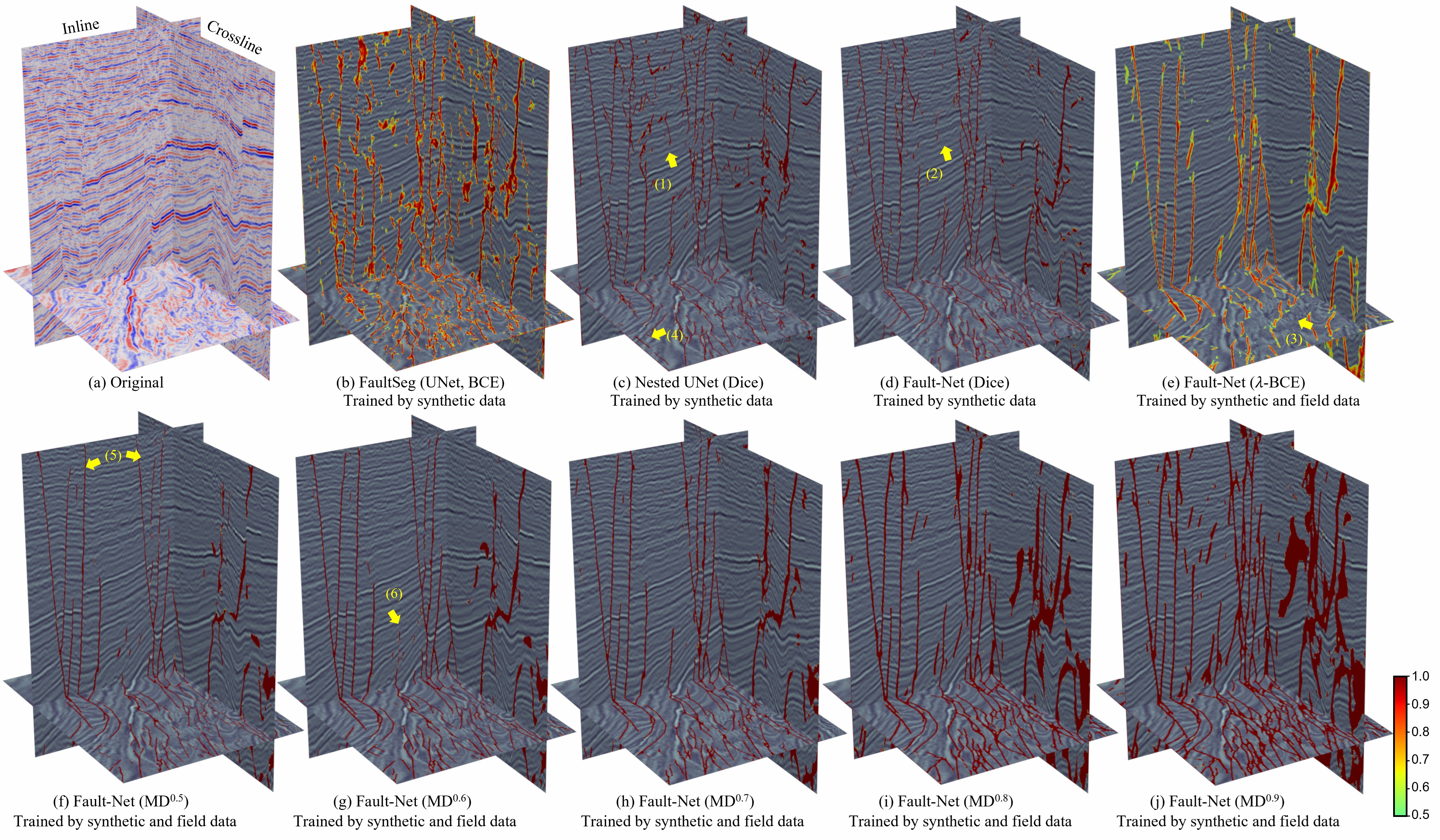}
	\centering\caption{Test results for Group-3 by work area $\mathcal{A}$. (b) is the result of FaultSeg, which was trained by synthetic data and BCE loss. (c) for Neted UNet, which was trained by synthetic data and Dice loss. (d) for Fault-Net trained with synthetic data. (e)-(j) using the mixture (Synthetic data, $\mathcal{B}, \mathcal{C}$ work area) dataset, and different loss functions, the variables for each result have been indicated in the figure.}
	\label{exp1}
\end{figure*}
\begin{figure*}[htb]
	\includegraphics[scale=0.65]{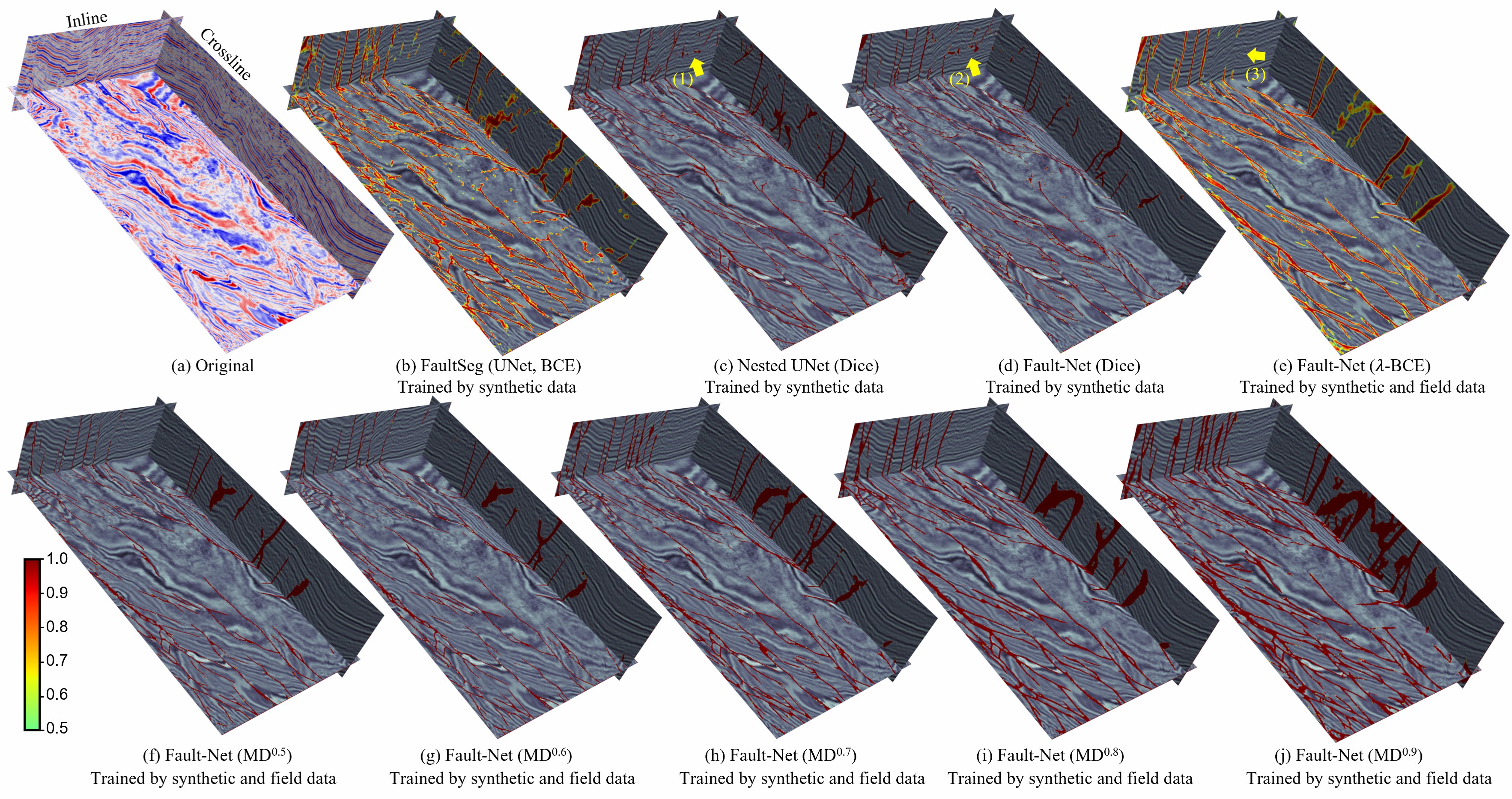}
	\centering\caption{Test results for Group-2 by work area $\mathcal{B}$. (b) is the result of FaultSeg, which was trained by synthetic data and BCE loss. (c) for Neted UNet, which was trained by synthetic data and Dice loss. (d) for Fault-Net trained with synthetic data. (e)-(j) using the mixture (Synthetic data, $\mathcal{A}, \mathcal{C}$ work area) dataset, and different loss functions, the variables for each result have been indicated in the figure. }
	\label{exp2}
\end{figure*}
\begin{figure*}[htb]
	\includegraphics[scale=0.55]{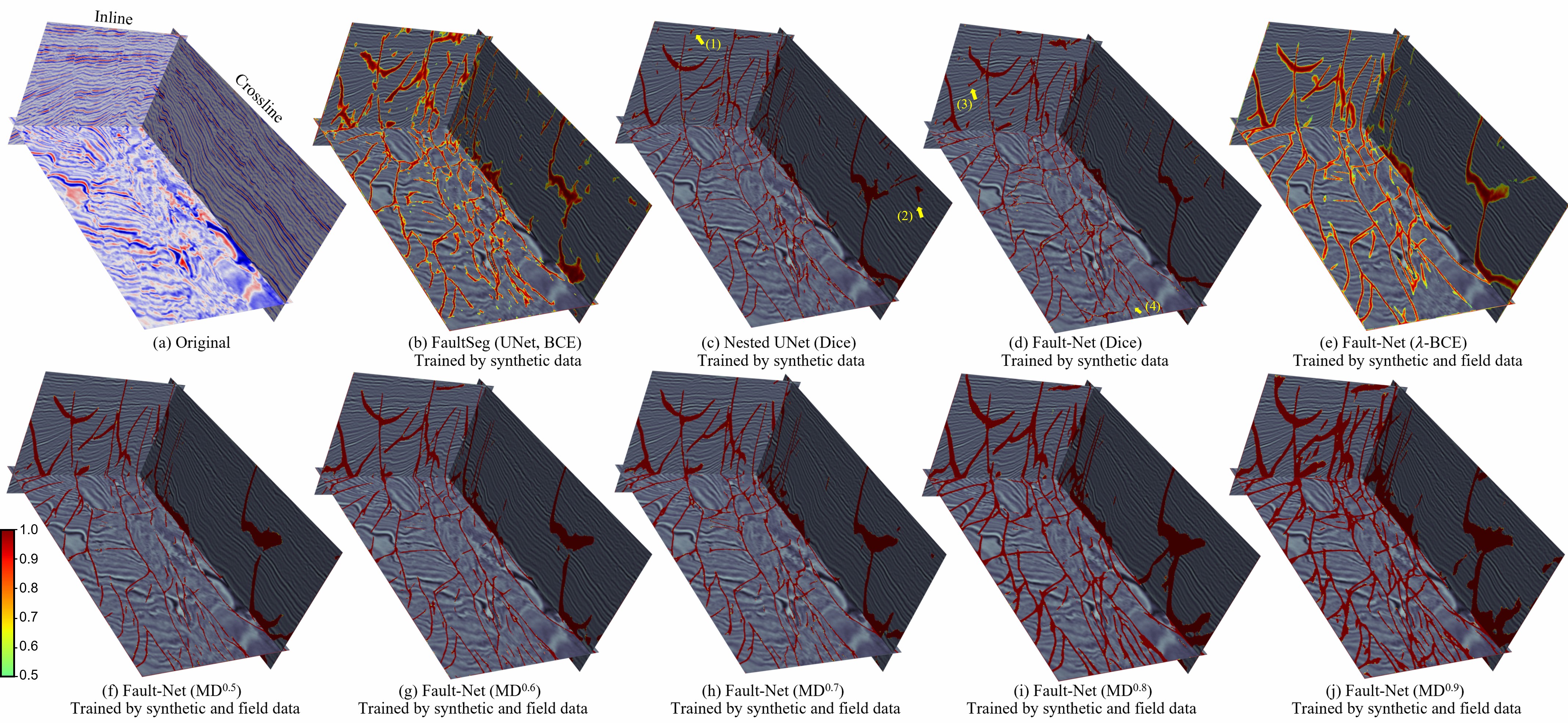}
	\centering\caption{Test results for Group-1 by work area $\mathcal{C}$. (b) is the result of FaultSeg, which was trained by synthetic data and BCE loss. (c) for Neted UNet, which was trained by synthetic data and Dice loss. (d) for Fault-Net trained with synthetic data. (e)-(j) using the mixture (Synthetic data, $\mathcal{A}, \mathcal{B}$ work area) dataset, and different loss functions, the variables for each result have been indicated in the figure. }
	\label{exp3}
\end{figure*}
In Table \ref{t3}, Fault-Net and Nested UNet have similar performance in terms of quantitative metrics, but in Table \ref{t1}, the former has only 3.4\% of the latter's computation. Fault-Net has stronger performance than UNet, but significantly lower computation than it. 
This well demonstrates the superiority of our network structure, which use of high-resolution characterization of the edge features for seismic faults can yield more promising results via fewer computational resources, as can be further illustrated by the qualitative experiments that follow. Moreover, experiments on synthetic data (Group 4) show the improvement of MCF for performance.

It should be noted that the network trained with synthetic data in Table  \ref{t3} seems to show the best performance, because the verification data also uses synthetic data, the training set and verification set have similar image features and distribution. The mixed training network adds field data. In the next qualitative experiment, we will show that the mixed data training model can obtain more reliable results (Fig. \ref{exp1}, \ref{exp2}, \ref{exp3}).

Fig. \ref{exp0} is calculated from Table \ref{t3}, which demonstrates the effect of FNL on various losses. As mentioned above, there are more FNLs in the annotation of work area $\mathcal{C}$ and thus may affect the training process of the model, so the overall performance of groups 2 and 3 is lower than that of Group-1. Fig. \ref{exp0} shows the performance difference between Group-1 and the other two groups, respectively, to illustrate the suppression effect of MD losses on FNLs. The figure indicates the MD loss compared to BCE has an suppression effect on FNL, and the difference between Group-1 and the control group gradually decreases as  $\gamma$  increases, bringing the difference to about 0 at  $\gamma=0.9$. However, in terms of performance, there is a significant decrease in the IOU metric when $\gamma$ is greater than 7. Therefore, we consider $\gamma=7$ as the more reliable parameter.

\subsubsection{Analysis of test set results.}

We use the optimal results of each group of experiments run on the corresponding test set to obtain the Fig. \ref{exp1}, \ref{exp2}, \ref{exp3}. We also add comparisons with current mainstream data-driven methods, such as those disclosed by Wu et al.\cite{wu2019faultseg3d} and Gao et al.\cite{gao2021fault}. Notably, these methods are trained using synthetic data, with Wu's method using its open source code\footnote{https://github.com/xinwucwp/faultSeg} and our work based on Gao's paper using Nested UNet to restore its training process and results as much as possible\footnote{Since there is no open source code for this work (Nested Residual UNet), we restore its network structure as much as possible based on Nested UNet (UNet++) according to the description in the paper. Nested UNet uses https://github.com/ShawnBIT/UNet-family.}.

Fig. \ref{exp1} shows the detection results for Group-3, which uses work area $\mathcal{A}$ as the test set, we have tagged each result with the network, loss and training data used. Among them, Fig \ref{exp1}-(b), (c), (d) are trained using synthetic data, and they demonstrate the difficulty of generalizing synthetic data on this work area. Although both (c) and (d) provide the explicit interpretation of the faults, their performance is still far from the same compared to other networks trained using mixed data, especially at Fig \ref{exp1}-(1), (2) which shows strong noise. Moreover, in practical applications, we would like to obtain more large continuous fault planes and fewer fault fragments, and the results provided by (f)-(j) are clearly more in line with this need.

Fig.\ref{exp1}-(c) Compared with (d), there are some obvious error detection caused by stitching (Fig \ref{exp1}-(4)). Because of the RAM limitation of Nested UNet (Table \ref{t1}), we cannot infer the whole 3D image input, but can only infer the result by cubing and then stitching, the process of stitching will lead to discontinuity of the result and error detection. The utilization of computational resources by Fault-Net is significantly better than other networks, which is one of the advantages of Fault-Net.

Fig. \ref{exp1}-(e) is trained using distribution-based loss because its equivalence of propagation gradients for foreground and background does not allow effective suppression of FNL, resulting in some significant false negatives of its model at (3).

Fig. \ref{exp1}-(f)-(j), as $\gamma$ increases, the "width" of detected faults gradually increases, and the network gradually tends to detect foregrounds (faults). When $\gamma$ is greater than 7, the network starts to show obvious false positives, which is the main reason why the IOU metric decreases at this time. And there are some underreporting at (5) and (6) in the figure, the inhibitory effect on FNL is not significant when $\gamma$ is less than 7.

Fig. \ref{exp2} shows the detection results for Group-2, which uses work area $\mathcal{B}$ as the test set. Compared to work area A, the network obtained by training only with synthetic data improves in the detection of work area $\mathcal{B}$, but still shows some false positives at (1) and (2). Similar to the detection results for Work area $\mathcal{A}$, Fault-Net trained using BCE loss shows some missed reports (at (3)). Similar conclusions can be obtained for the results of Fig. \ref{exp2}-(f)-(j) in work area $\mathcal{B}$ as in work area $\mathcal{A}$, i.e., the increase of the $\gamma$ makes the network more inclined to predict as foreground (faults). Among these results, our method (h) works the most effective. Furthermore, our results are much smoother than those using exclusively synthetic data, especially for the consistent description of large faults.

Fig. \ref{exp3} shows the detection results for Group-1, which uses work area $\mathcal{C}$ as the test set. In this work area, compared to the network trained with the mixed data set, the network trained with only synthetic data still does not generalize consistently, and as mentioned above, although it demonstrates better quantitative metrics in the validation set, its performance in the three test sets is still inferior to that of the network trained with the mixed data. This demonstrates the significant superiority of MD loss, unlike the general loss function, which supports the inclusion of human experience (few labeled slices) in the training and can suppress the FNL in it, giving more promising results on work area that cannot be generalized by synthetic data, such as the obvious error detection at Fig. \ref{exp3}-(1)-(4). Moreover, these experiments also show a common feature that the region-based loss (Dice, MD) gives more stable fault detection results with a judgment probability close to $1.0$ for faults, while the distribution-based loss (BCE, $\lambda$-BCE) gives a probability between $[0.5-1.0]$. Thus MD loss gives better results than $\lambda$-BCE in the detection of work area $\mathcal{C}$, even if most of the manual labeling is correct, and MD loss is also the first region-based loss function that can be trained using sparse labels.

\subsection{Testing on public data}
In this subsection we compare the performance of these methods on publicly available data, where our method uses only the network obtained from MD$^{0.7}$ training in Group-1. This subsection of the experiment incorporates a comparison with AAM-UNet \cite{dou2021seismic}, which uses active attention to improve UNet and is trained by $\lambda$-BCE and mixed datasets.

\begin{figure*}[htb]
	\includegraphics[scale=0.63]{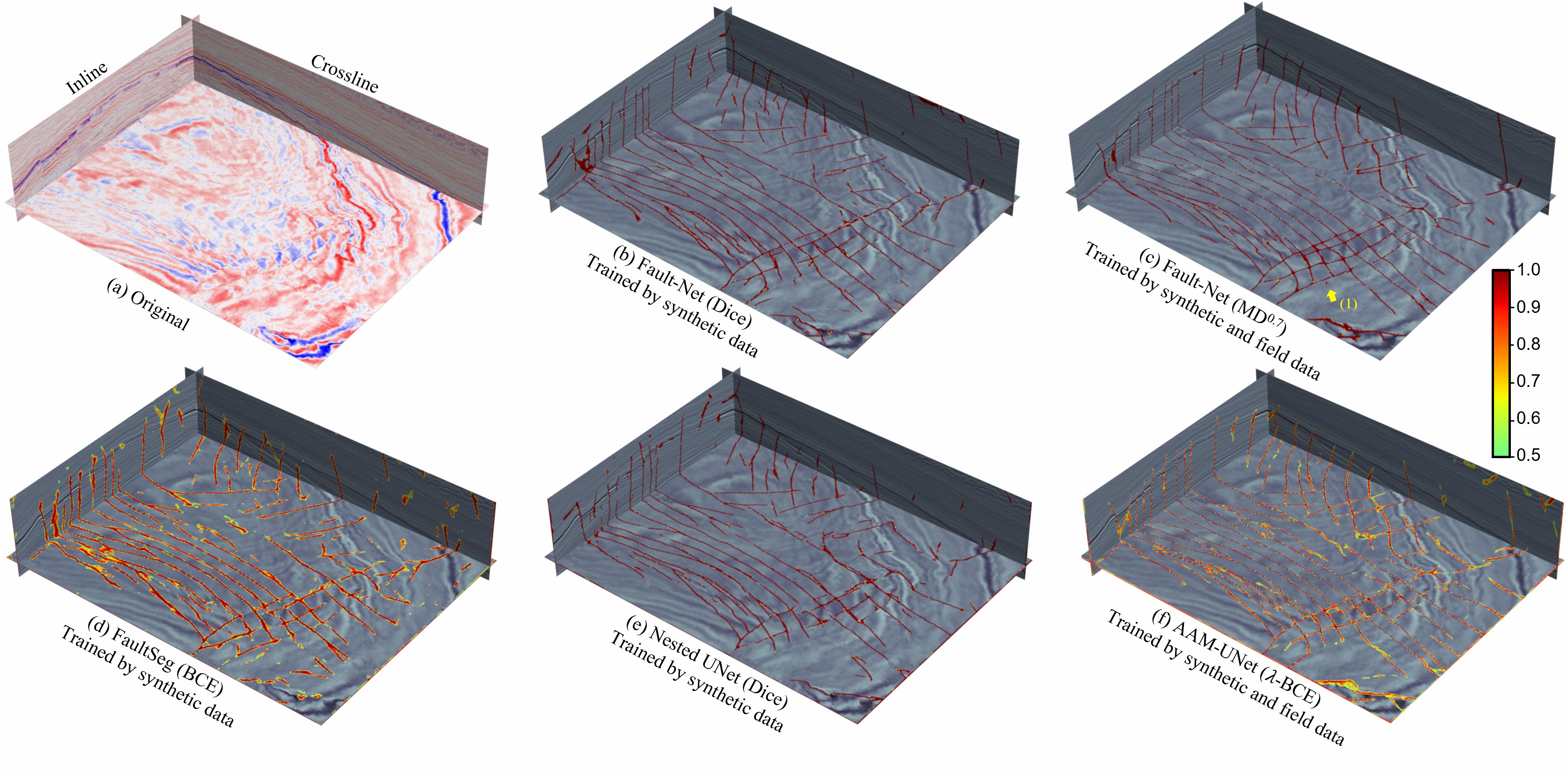}
	\centering\caption{Tests on Netherlands F3. The experimental variables for each result are indicated.}
	\label{exp4}
\end{figure*}

\subsubsection{Netherlands F3}
First we tested on the Netherlands offshore F3 data (Fig. \ref{exp4}), in which we intercepted a more fault-rich region where there are some criss-cross faults, the selected region is sampled at a total of $128 \times 512 \times 384$ grid points. Most of the methods can do well to detect faults in the vertical inline direction among them, while for faults parallel to the inline they show discontinuities and significant misses. In Fig. \ref{exp5}, only our method (Fig \ref{exp4}-(c)) has a clear and unambiguous detection of the fault present at (1), where the method demonstrates a great advantage. 

Fig. \ref{exp4}-(f) has less noise compared to (d), but as mentioned above, the distribution-based loss is difficult to provide a stable interpretation of the fault, and it gives probabilistic and ambiguous results.

Furthermore, Fault-Net demonstrates similar performance to Nested UNet when without MD loss (without adding field data training), while the dependence on computational resources between the two is far from each other. Fig. \ref{fig4}-(d-f) shows some obvious discontinuities and traces caused by splicing, while the results of (b) and (c) are smoother and continuous. our method supports inference on larger size seismic data in the same size of memory, which can avoid many problems caused by splicing.

\subsubsection{New Zealand Kerry}
The data is publicly available on the SEG Wiki, and we intercepted the fault-rich regions of it, the selected region is sampled at a total of $272\times 608 \times 192$ grid points.

\begin{figure*}[htb]
	\includegraphics[scale=0.65]{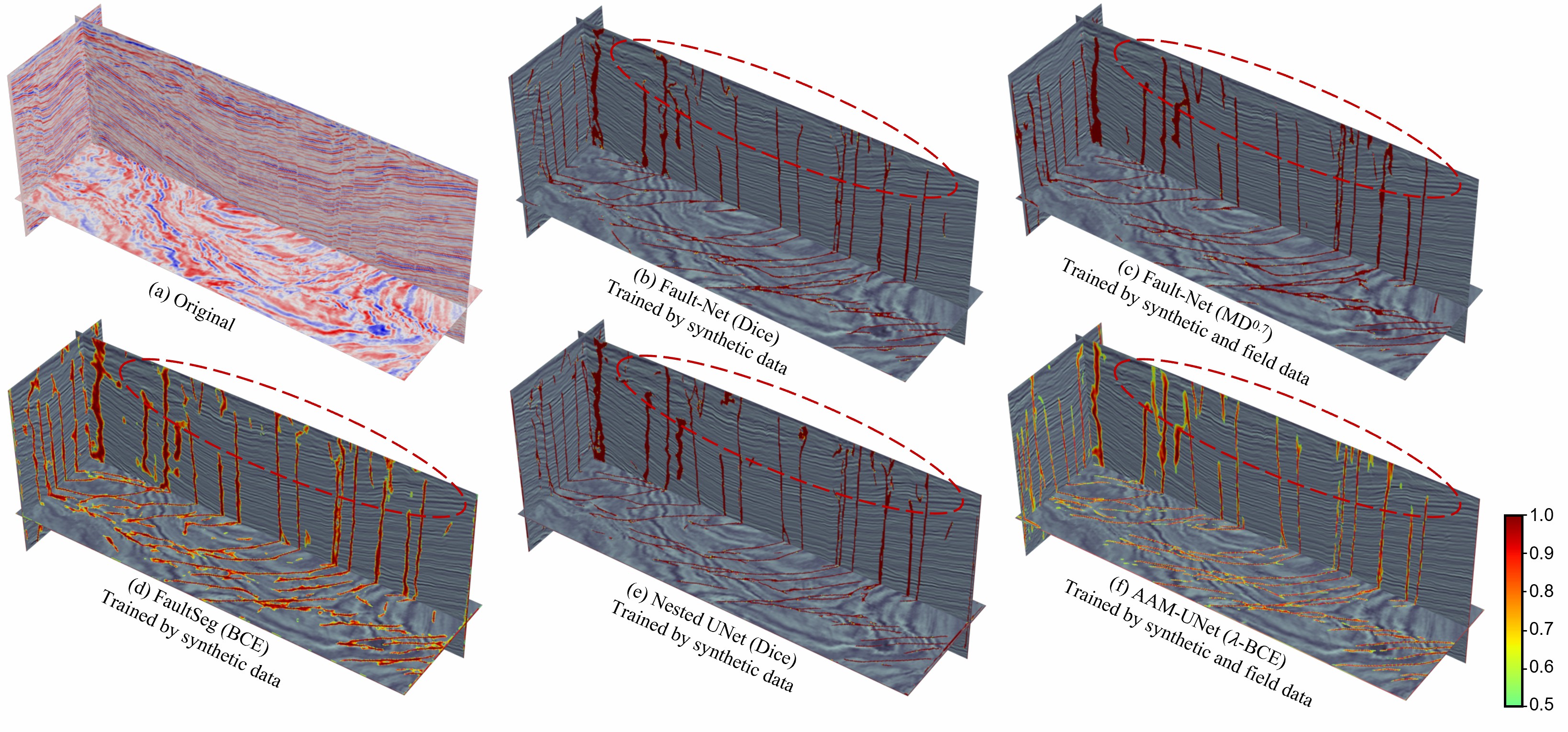}
	\centering\caption{Tests on New Zealand Kerry. The experimental variables for each result are indicated.}
	\label{exp5}
\end{figure*}

There are a great number of vertical faults in the region, and the faults are more obvious in the reflector, especially the fault development and geological structure of the region are similar to the synthetic data released by Wu \cite{wu2019faultseg3d}, so all the networks in Fig. \ref{exp5} show more reliable detection results. Even so, our method has an advantage in terms of the details of fault detection.

In Fig. \ref{exp5}, the network ((c) and (f)) that incorporates the field data training gives clear detection results with almost no noise. As for the other networks, because of the irregular reflection at the top of the data (red circle), the network did not learn the corresponding processing in the synthetic data, leading to some noisy and wrong detections, and (c) has a more stable and clear fault interpretation compared to (f).

\begin{figure*}[htb]
	\includegraphics[scale=0.63]{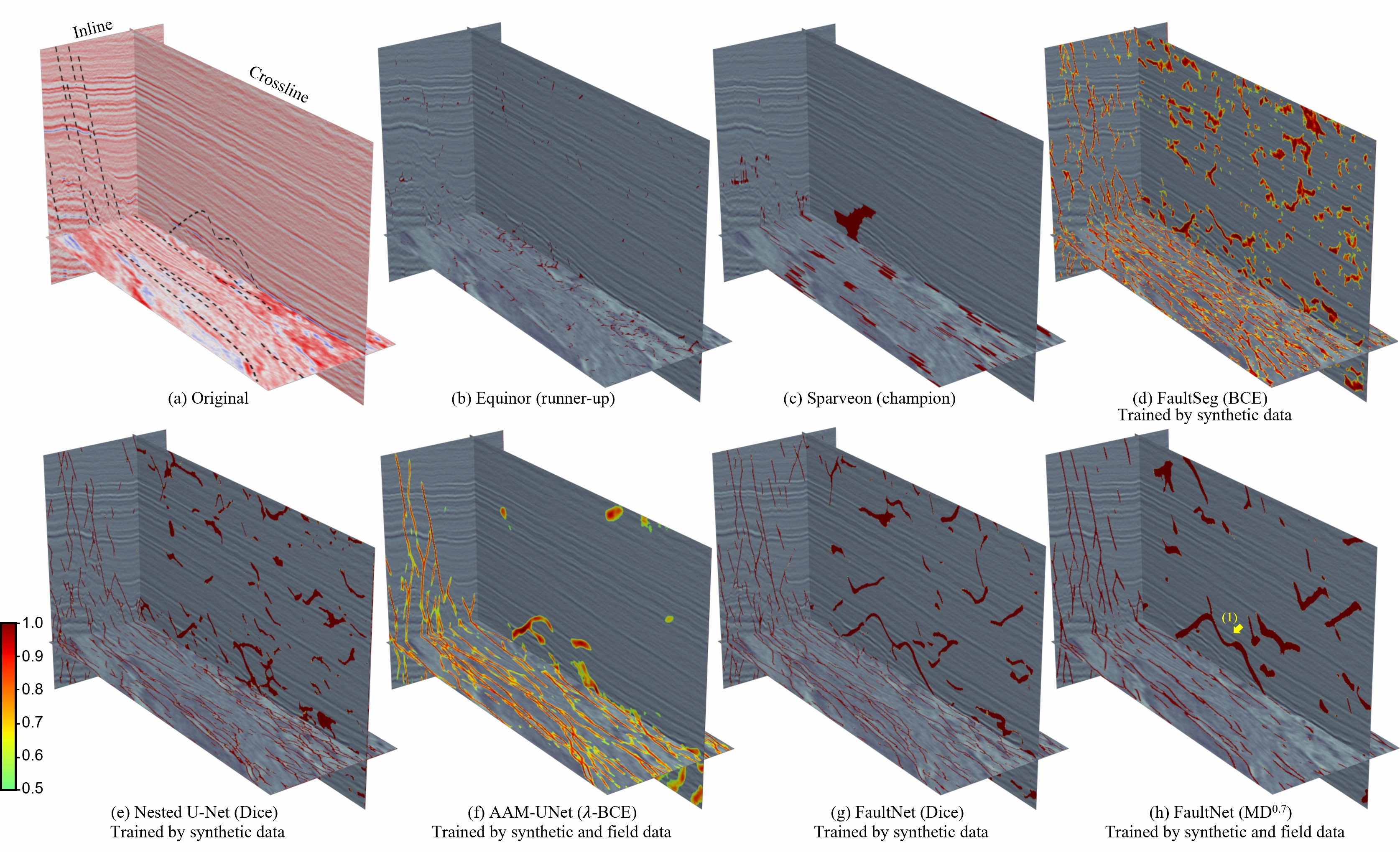}
	\centering\caption{Tests on FORCE ML Competition Adele. The experimental variables for each result are indicated.}
	\label{exp6}
\end{figure*}

\subsubsection{FORCE ML Competition}
This data was provided by FORCE ML Competition 2020. This competition wanted to find out if these very efficient modern machine learning based fault detection algorithms perform equally as good on not so perfect data. The competition had 80 teams registered, the final results of this competition are published in this URL\footnote{FORCE ML Competition, https://drive.google.com/drive/folders/\ 1Hu4VJN9xLOWixSMdf2xN6fRk0zmOz-1J}, as well as a demonstration of each team’s testing results.

The competition provided a validation seismic image from the Ichthys Field on the NW Shelf of Australia, the blind dataset that organizers provided to the contestants comes from the Adele seismic survey that is located some 15 to 20 km to the NE of the Ichthys seismic survey, the selected region is sampled at a total of $512\times 288 \times 768$ grid points. As stated by the organizers, the faults in the seismic images of this region are not perfect and difficult to interpret even by humans, but our method still gives very satisfactory results.

Because this survey fault is difficult to observe, we have included some manual interpretation in Fig. \ref{exp6}-(a) to annotate the main faults in it, which researchers can also compare by downloading the original data. The winners (Sparveon and Equinor) both demonstrated poor detection, and the competition results showed that detecting faults in this survey is very challenging.

The similarity of the wave number spectrum of this data noise to that of the faults leads to the overdetection of FaultSeg and Nested UNet (Fig. \ref{exp6}-(d) and (e)) on crossline and timeline, and the (f) and (g) timeline also shows some noise, these fault-like noise detections are not geologically credible, but (e), (f) and (g) all identify the major faults in inline, while the faults in (d) inline show discontinuities and substantial omissions. Of these methods (h) shows the best performance, not only detecting the main faults in the inline, but also demonstrating high signal-to-noise ratios in the other two directions, with stronger geological interpretation. Furthermore, our method detects more straight compared to other results, and is therefore more geologically reasonable and easier to interpret, curved faults can be more difficult to relate  with conventional geologic evolution interpretation than straight faults. Furthermore, our method detects more large fault planes and relatively fewer small fault fragments, unlike the method trained using synthetic data exclusively, which clearly has more fault fragments and noise in (d), (e) and (g).

\section{Conclusion}
We present MD loss, which supports the incorporation of human experience through a few 2D slices in the training of the 3D network's faults and suppresses abnormal annotations in the manual labels, thus significantly improving the model performance and generalizes it to more surveys. Meanwhile, Fault-Net was developed, which uses high-resolution representational features and MCF to ensure that the edge features of faults are fully preserved during inference, allowing more reliable results with significantly lower computational resources than existing networks.

We illustrate the characteristics and effectiveness of MD loss using three surveys and synthetic data from Sinopec. Subsequently, in combination with the three publicly available data, it is verified that the network trained by adding field data through MD loss outperforms the network without it. MD loss can compensate for the limitations of the currently widely used synthetic-data-based methods.
With Dice loss, Fault-Net shows equal or better performance than Nested UNet, but it requires only 3.4\% of the computational resources of Nested UNet, and its comprehensive performance is the best among several major fault detection networks.
Our method can help extend fault detection capabilities for unknown surveys by adding human experience, and provide faster and more reliable fault detection results.
% use section* for acknowledgment
\section*{Acknowledgment}
The authors are very indebted to the anonymous referees for their critical comments and suggestions for the improvement of this paper. This work provides open test script:  https://github.com/douyimin/FaultNet.

\bibliographystyle{IEEEtran}
\bibliography{references}

\end{document}